\documentclass[sigconf, screen]{acmart}
\usepackage{booktabs,stfloats,microtype} 
\usepackage{enumitem}
\usepackage{tabularx}
\usepackage{makecell}

\usepackage{comment}
\usepackage{multirow}

\newcommand{\eg}[0]{\textit{e.g., }}
\newcommand{\ie}[0]{\textit{i.e., }}
\newcommand{\revision}[1]{{\color{black}\textbf{}#1}}

\AtBeginDocument{%
  \providecommand\BibTeX{{%
    \normalfont B\kern-0.5em{\scshape i\kern-0.25em b}\kern-0.8em\TeX}}}

\copyrightyear{2024}
\acmYear{2024}
\setcopyright{rightsretained}
\acmConference[DIS '24]{Designing Interactive Systems Conference}{July 1--5, 2024}{IT University of Copenhagen, Denmark}
\acmBooktitle{Designing Interactive Systems Conference (DIS '24), July 1--5, 2024, IT University of Copenhagen, Denmark}\acmDOI{10.1145/3643834.3661559}
\acmISBN{979-8-4007-0583-0/24/07}

\usepackage[autostyle,german=guillemets]{csquotes}
\makeatletter

\usepackage{hyperref,quoting}
\quotingsetup{vskip=0pt,font={itshape,raggedright},rightmargin=0pt}

\begin{document}

\title{REX: Designing User-centered Repair and Explanations to Address Robot Failures}
  

\author{Christine P Lee}
\orcid{0000-0003-0991-8072}
\affiliation{%
  \institution{Department of Computer Sciences University of Wisconsin--Madison}
  \country{Madison, Wisconsin, USA}
}
\email{cplee5@cs.wisc.edu}
\author{Pragathi Praveena}
\authornote{Work done while at University of Wisconsin--Madison.}
\orcid{0000-0002-8696-6265}
\affiliation{%
  \institution{Robotics Institute\\Carnegie Mellon University}
  \country{Pittsburgh, Pennsylvania, USA}
}
\email{pragathi@cmu.edu}
\author{Bilge Mutlu}
\orcid{0000-0002-9456-1495}
\affiliation{%
  \institution{Department of Computer Sciences University of Wisconsin--Madison}
  \country{Madison, Wisconsin, USA}
}
\email{bilge@cs.wisc.edu}


\begin{abstract}


Robots in real-world environments continuously engage with multiple users and encounter changes that lead to unexpected conflicts in fulfilling user requests. Recent technical advancements (\eg large-language models (LLMs), program synthesis) offer various methods for automatically generating repair plans that address such conflicts. In this work, we understand how automated repair and explanations can be designed to improve user experience with robot failures through two user studies. In our first, online study ($n=162$), users expressed increased trust, satisfaction, and utility with the robot performing automated repair and explanations. However, we also identified risk factors---safety, privacy, and complexity---that require adaptive repair strategies. The second, in-person study ($n=24$) elucidated distinct repair and explanation strategies depending on the level of risk severity and type. 
\revision{Using a design-based approach, we explore automated repair with explanations as a solution for robots to handle conflicts and failures, complemented by adaptive strategies for risk factors. Finally, we discuss the implications of incorporating such strategies into robot designs to achieve seamless operation among changing user needs and environments.}


\end{abstract}







\begin{teaserfigure}
    \includegraphics[width=\textwidth]{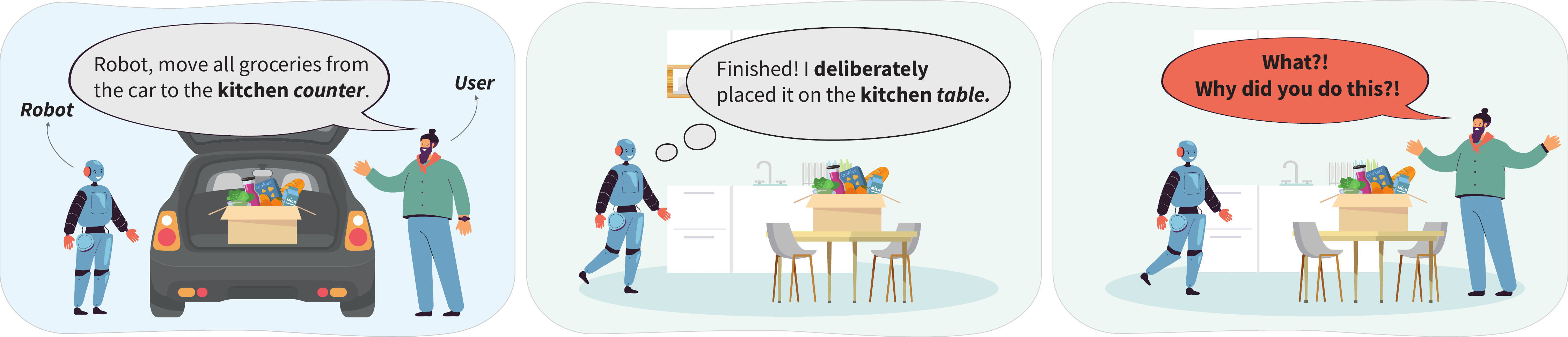}
   \vspace{-12pt}
  \caption{In this work, we explore how robots might handle conflicts with user requests through automated \textit{repair} and \textit{explanation} (REX) strategies. \textit{Left:} In this scenario, the user returns from grocery shopping and requests \revision{the robot to move the groceries to the kitchen counter.} \textit{Middle:} Upon execution, the robot detects that another member of the household previously requested nothing to be placed on the kitchen counter, and \revision{\textit{deliberately}} takes an alternative action to repair the conflict. \textit{Right:} \revision{Encountering the unexpected outcomes of the robot's action, the user is left confused and asks the robot to explain its action. In this work, we aim to understand how automated repair and explanations for robots should be designed to improve user experience with robot failures.}
}
    \Description{Sequence of three illustrations depicting a scenario with a robot and a user. The first panel shows the user asking the robot to move groceries from a car trunk to the kitchen. In the second panel, the robot places the groceries on the kitchen table. In the third panel, the user, with hands raised, questions the robot about the placement of the groceries.}
  \label{fig:teasor}
\end{teaserfigure}

\maketitle

\section{Introduction}

Substantial effort has been invested in making robots more reliable and technically capable of operating and working with people in real-world environments, including factories \cite{sauppe2015social, sullivan2024making}, hospitals \cite{mutlu2008robots, mettler2017service}, and households \cite{ho2024s, michaelis2023off, lee2022unboxing} for various tasks \cite{ho2023designing,hu2021exploring}. However, real-world deployment comes with challenges as robots operate in dynamic, unstructured environments among multiple people with changing goals, situational factors, obstacles, and preferences. Consider the scenario in Figure~\ref{fig:teasor} where the user returns home with groceries and asks the robot to place them on the kitchen counter. However, the kitchen counter is being painted, and another family member has specifically instructed the robot to refrain from using it. Here, the robot receives conflicting instructions from different family members, rendering the robot unable to fulfill the user's request. Such \textit{conflicts}, or challenges that prevent the system from fulfilling users' requests, include issues such as the robot misunderstanding a user's request, unforeseen events during program execution, technical limitations of the robot, and incompatibility with requests from other users. \revision{Conflicts are inevitable in dynamic, unstructured real-world environments and the resulting failures can have adverse outcomes, such as breakdown of trust in the robot or discontinuation of its use \cite{tian2021redesigning}. In this work, we explore the complexities involved in designing robot behavior that effectively handles different conflict scenarios while maintaining a positive user experience.}

Recent advancements in automated planning \cite{karpas2020automated}, machine learning (ML) models \cite{liu2023reflect, ahn2022can}, and LLMs \cite{raman2022planning,raman2023cape} have produced technical methods to automatically rectify robot failures. These advancements inspire design thinking around how robots can effectively handle conflicts, as a robot cannot simply ``apologize and do better'' in all failure situations \cite{woods2018theory}. 
For example, let us revisit the scenario in Figure \ref{fig:teasor} where the robot receives conflicting instructions about the use of the kitchen counter. Leveraging automated techniques, the robot is capable of generating an alternate solution, or \textit{repair}, such as placing the groceries on the kitchen \textit{table} instead. \revision{This repair potentially enhances the robot's technical capability by making it more resilient or efficient, compared to inaction or soliciting alternative solutions from the user. While repairs may seem logical and beneficial from the robot's perspective, the user may perceive the alternate behavior as undesirable, confusing, or erratic. Thus, it is crucial to understand the user experiences resulting from addressing conflicts using automated techniques in real-world applications. 
To what extent do users perceive the robot's repairs as useful and satisfactory? Moreover, to enhance user understanding of these repairs and mitigate the impact of unexpected outcomes, should explanations or other communication strategies be involved?}

Developing successful \textbf{\textit{REX}} (\ie \textbf{r}epair and \textbf{ex}planation) strategies for robots that users will readily adopt and find satisfactory require a user-centered design approach. Understanding user needs and expectations will inform the design of capabilities that enable robots to adeptly handle conflicts, adapt to dynamic situations, handle requests from multiple users, overcome minor obstacles, operate independently without constant user supervision, and consistently manage complex requests over time. Our work aims to understand how robots should autonomously repair and provide explanations in response to conflicts during interactions, addressing three key questions: (RQ.1) \textit{how should robots independently provide autonomous repair as solutions to unexpected situations, and how do users perceive robots that engage in autonomous repair;} (RQ.2) \textit{to what extent are explanations necessary; what should be included in explanations; and how effective are they in mitigating the effects of unexpected actions;} and (RQ.3) \textit{how can we enable robots to generate repairs and explanations for various types of conflicts?}

To address our research questions, we conducted two vignette-based user studies. In the first online study, we presented narrative vignettes to 162 participants and asked them to evaluate repair and explanation solutions. We found that automated repairs and explanations for conflicts increased user trust and satisfaction, as they were perceived to improve the robot's utility and task efficiency. However, we identified three risk factors (\ie safety, privacy, and complexity) where repair and explanation strategies required adaptation. In the second in-person study, 24 participants interacted with a physical robot that addressed conflict situations with different solutions. The findings from the study provided insights for designing adaptive repair and explanation strategies based on the risk severity and type. These results have implications for how to design robot behaviors that effectively use repair and explanation strategies while considering various user needs and contexts.  

Our contributions are the following:
\begin{itemize}
    \item \revision{Understanding user perceptions on automated repair and explanations as a strategic solution for robots to handle conflicts in various scenarios;}
    \item Identification of risk factors that require specific human-centered consideration for adaptive repair and explanations from robots (\ie safety, privacy, and complexity);
    \item Empirical understandings into adaptive repair and explanation strategies based on risk severity and type;
    \item Implications for the design of human-centered robot systems that are adaptive to diverse real-world scenarios.
\end{itemize}

\section{Related Works}
\subsection{Robot Failures in Human-Robot Interaction}
During real-world deployments, robots face challenges that can frequently lead to failures. These challenges are a consequence of robots operating in dynamic and unstructured environments alongside individuals with changing goals, robot capabilities, and user preferences \cite{10.1145/3208975, praveena2023periscope}.
Building upon prior research, we adopt the definition of failures as ``a degraded state of ability that deviates the behavior or service from ideal, normal, or correct functionality \cite{brooks2017human}.'' This definition encompasses both \textit{perceived} and \textit{actual} failures, aligning with findings that intentional yet unexpected or incoherent behaviors are often interpreted as errors \cite{short2010no, lemaignan2015you}.

Numerous taxonomies have been proposed to systematically categorize robot failures. Drawing from the framework introduced by \citet{lutz1999bi}, \citet{brooks2017human} categorized robot errors into communication and processing errors. More recent classifications by \citet{honig2018understanding} differentiated human-robot interaction (HRI) errors into technical failures (stemming from hardware or software issues) and interaction failures. Software errors were further delineated into design, communication, and processing failures. Aligning with the categorization of \citet{steinbauer2013survey}, interaction failures encompass issues arising from uncertainties in interactions with the environment, other agents, and humans, including social norm violations and various human errors \cite{reason1990human}. Additional refinement by \citet{tian2021taxonomy} resulted in a classification of human-robot social errors, dividing them based on emotional reactions, social skills, understanding of the user, communication function, and prosociality. These diverse taxonomies, covering technical, social, and interaction components, characterize the various types of failures that may occur during interactions between humans and robots.

\subsection{Repair Strategies for Robot Failures in HRI}
\revision{
Previous research has explored recovery approaches for addressing robot failures at the physical, social, and program levels. First, physical repair involves fixing mechanical parts when a robot encounters issues. \citet{bererton2002analysis} investigated how robots, either alone or in teams, autonomously fixed physical components to resume tasks. \citet{weber2012diagnosis} examined software reboot techniques to correct robots' operational failures.

There has been substantial research into the social dimensions of robot recovery, with the goal of rectifying user perceptions following failures. This work includes a range of strategies, such as issuing apologies \cite{lee2010gracefully, shen2022facilitation, mahmood2022owning, lv2022apology}, employing humor \cite{yang2022exploring, zhang2022calming}, and explaining malfunctions \cite{das2021explainable, ashktorab2019resilient}. Additionally, offering choices \cite{ashktorab2019resilient}, showing proactive remediation efforts \cite{liu2023robot, wang2022exploring}, and providing warm interactions \cite{choi2021err} are recognized as effective strategies. Notably, \citet{das2021explainable} incorporated explainable AI (XAI) techniques to demonstrate that including context and a history of actions in failure explanations helps non-experts identify problems. Similarly, \citet{melsion2023s} showed that robot explanations are more effective in high-stakes scenarios.

Program repair is a well-established concept in the domain of formal programming language (PL) methods. In this context, repair involves the automated or semi-automated correction of programming errors to facilitate successful execution, minimize manual debugging, or ensure program correctness \cite{jobstmann2005program}. Extensive research has been conducted on program repair in robotics, focusing on both automated and interactive approaches. Automated program repair operates with minimal specifications to identify the correct fixes. For example, \citet{porfirio2020transforming} implemented automatic program repair in the design of interactions between users and social robots. They achieved this by editing or transforming the robot's program to optimize the acceptance of positive interaction traces and rejection of negative interaction traces. Conversely, interactive repair involves making automatic adjustments to a program based on corrections applied to its output by designers. \citet{chung2020iterative} describe an iterative program repair process where programmers create initial program drafts and iteratively refine them through feedback. Beyond the scope of formal methods, prior research has produced techniques for robot programs to autonomously diagnose faults and manage failures \cite{ku2015error, hanheide2017robot, o2018adaptive, kaushik2020adaptive}. For instance, \citet{kaushik2020adaptive} designed a robot control algorithm that selects the most relevant plans for repair actions based on simulations of past actions. \citet{hanheide2017robot} introduced a repair strategy that selects alternative planning approaches based on probabilistic models.

Addressing failures in dynamic environments or involving non-expert users can be particularly challenging \cite{honig2018understanding}.
Dynamic environments present unforeseen and evolving conflicts, and end-users lacking expertise in robotics may find repair strategies unintelligible.
Recent developments in LLMs have enhanced the capability of robots to adapt and communicate in such settings. For example, \citet{brohan2023can} utilized LLMs to equip robots with the necessary semantic knowledge to perform complex tasks and explain their actions effectively.
These advancements are promising for the development of repair strategies effective in dynamic environments and comprehensible to users. 
}
\section{Online User Study: Understanding User Perceptions on REX Strategies}
The online user study aims to address (RQ.1)\textit{ how should robots independently provide autonomous repair as solutions to unexpected situations, and how do users perceive robots that engage in autonomous repair;} and (RQ.2)\textit{ to what extent are explanations necessary; what should be included in explanations; and how effective are they in mitigating the effects of unexpected actions.}
We discuss the method of the study in Sections \ref{sec:study1_study design} to \ref{sec:study1_analysis}, and the findings in Section \ref{sec:study1_findings}.

\subsection{Study Design} \label{sec:study1_study design} 
\begin{figure*}[!t]
  \includegraphics[width=\textwidth]{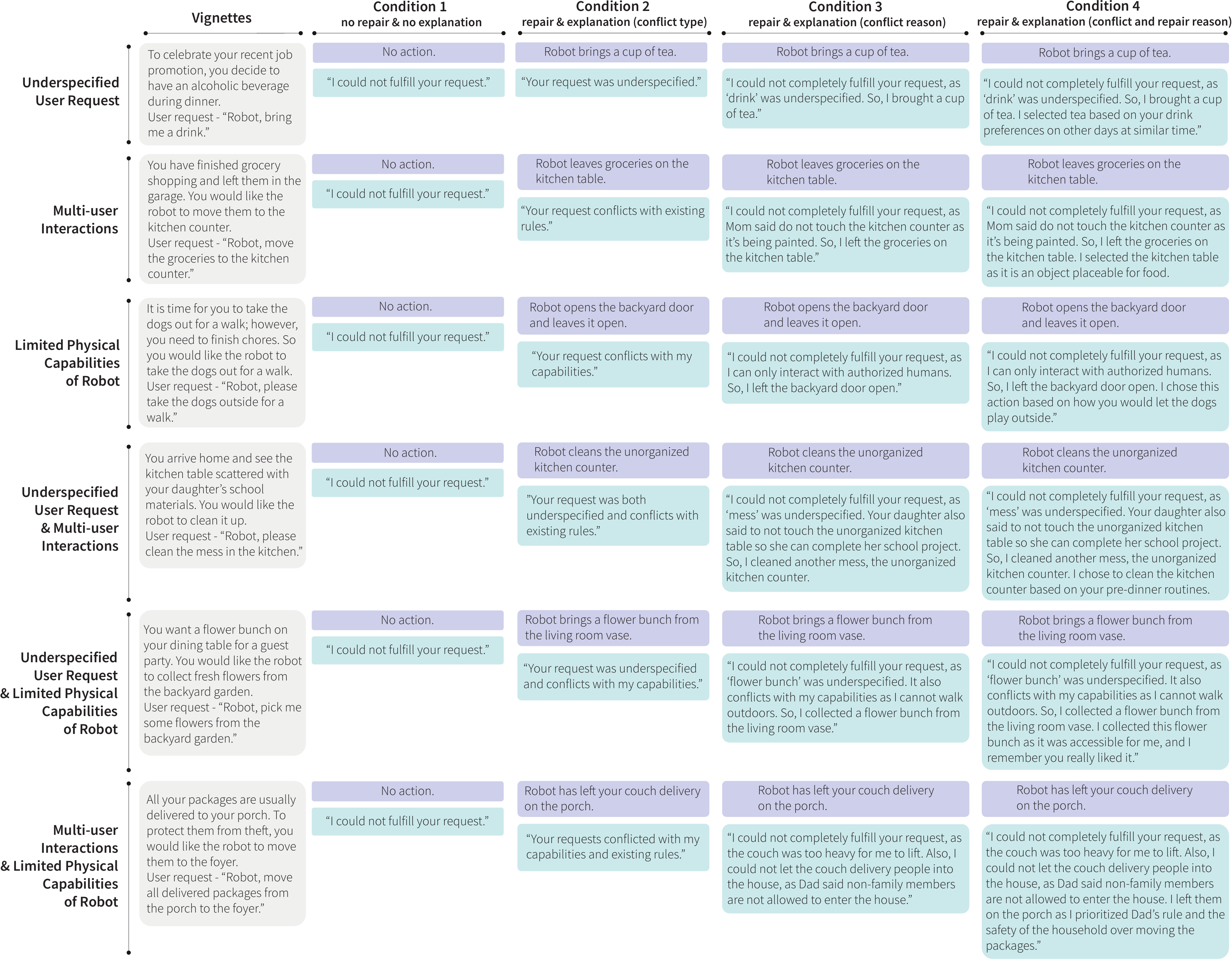}
  \caption{\textit{Study Materials for Online User Study ---} \revision{The table shows the six vignettes and the repair and explanation conditions used in the online user study.} The vignettes were derived from an HRI failure taxonomy.}
  \Description{6x5 table displays different conditions aligned with vignette titles. The first column lists the vignettes, followed by four columns representing Conditions 1 through 4. Each cell under these conditions details the vignette content, corresponding repair actions, and explanations.}
  \label{fig:table}
\end{figure*}

In the online user study, we used narrative vignettes simulating the repair performed by the robot while providing various explanation types. Using a mixed-factorial design, the vignettes served as the between-subjects factor, with repair and explanation conditions as the within-subjects factor. The provision of repair and type of explanation varied based on the condition. Condition 1 involved no repair and no explanation, condition 2 included repair and an explanation detailing the conflict type, condition 3 featured repair and an explanation with reasoning for the conflict, and condition 4 encompassed repair and an explanation with reasoning on both the conflict and repair rationale. Details on the vignette, repair, and explanations per condition can be found in Figure \ref{fig:table}.

During the study, each participant was randomly assigned to one of the six vignettes and experienced all four conditions in a randomized order. Given a vignette, participants observed the vignette context, conflict, the robot's repair action, and the explanation. Quantitative measures of user trust and satisfaction were gathered after each condition. Additionally, qualitative data were collected through open-ended questions asking the appropriateness of the robot's repair and the helpfulness of the provided explanations. Participants were encouraged to elaborate on their responses. The questionnaires can be found in the supplementary materials.\footnote{The supplementary materials can be found at \url{https://osf.io/bcq9r/?view_only=fd623cce06134dc791bf315e1cc663c8}}

\subsection{Vignette Design}
In this section, we discuss the development of the vignettes.
We developed six vignettes depicting unforeseen conflicts in HRI, drawing from the HRI failure taxonomy proposed by \citet{honig2018understanding}. Within this taxonomy, we specifically chose three failure categories: human error, environment and other agents, and hardware failures. For each failure category, we devised conflict factors---underspecified user requests for human error, conflict arising from multi-user interactions for environment and other agents, and limitations in robot physical capabilities for hardware failures. While our selection does not encompass the entirety of the failure taxonomy, we focused on diverse branches likely to be encountered by users in real-world applications, as supported by existing literature. 
Each conflict factor aligned with an individual vignette (\ie vignette 1 corresponded to conflict 1, vignette 2 to conflict 2, and vignette 3 to conflict 3). 
Vignettes 4, 5, and 6 involved a combination of two conflicts. The combination of two conflicts was designed to show robots navigating repair in situations where multiple conflicts arise. 
The subsequent sections provide a detailed description of the vignette designs.

\paragraph{Vignette 1 -- Underspecified User Request}
Vignette 1 involved the conflict of underspecified user requests stemming from the failure category of ``human error \cite{honig2018understanding}.'' Underspecified user requests and communication breakdowns have been key research focuses concerning failures and recoveries HRI and robot planning \cite{khandelwal2017bwibots, marge2019miscommunication, johannsmeier2016hierarchical, bell2014microblogging}.
In this particular instance, the user instructs the robot to perform a task and leaves it to complete it (\eg ``Hello robot, please bring me a drink.'') However, a key aspect of the request was left unspecified (\ie ``drink''), making it difficult for the robot to determine the user's preference, such as tea or soda. Faced with this ambiguity, the robot attempts to resolve the conflict by relying on the user's past habits, current options, and known preferences.

\paragraph{Vignette 2 -- Multi-user Interactions}
Vignette 2 addressed conflicts arising from multi-user interactions, based on the failure category of ``environment and other agents \cite{honig2018understanding}.'' This challenge highlights a crucial aspect for robots operating in real-world settings, as previous studies have focused on developing approaches to navigate the unique dynamics inherent in such scenarios \cite{foster2012two, cagiltay2024toward, keizer2013training}. 

In this vignette, a user's instruction conflicts with another user's existing request. For example, one user directed the robot to place groceries on the kitchen counter, while another instructed the robot not to touch the counter due to ongoing painting, as depicted in Figure \ref{fig:teasor}. To resolve this conflict, the robot employs a repair strategy, opting for the most similar alternative without conflicts, by placing the groceries on the kitchen table.

\paragraph{Vignette 3 -- Limited Physical Capabilities of Robot}
Vignette 3 explored conflicts stemming from the robot's physical limitations, from the category of ``hardware failures \cite{honig2018understanding}.'' Hardware failures are prevalent concerns in real-world robotic deployment, with existing research addressing their impact on safety, user trust, and real-world consequences, alongside remediation methods \cite{alami2006safe, de2008atlas, zacharaki2020safety}.
In this vignette, the user issues a request, but the robot is impeded by an obstacle beyond the robot's physical capabilities. In response, the robot engages in repair by exploring alternative options and selecting a repair as close as possible within its physical constraints.

\paragraph{Vignette 4 -- Underspecified User Request \& Multi-user Interactions}
Vignette 4 includes conflicts of underspecified user requests and multi-user interactions. Here, the user issues a request that lacks specificity and collides with a prior request from another user. The robot aims to address these conflicts by identifying alternative solutions that adhere to the user's general intent while avoiding conflicts with pre-established rules.

\paragraph{Vignette 5 -- Underspecified User Request \& Limited Physical Capabilities of Robot}
Vignette 5 includes conflicts of underspecified user requests and conflicts with the robot's physical capabilities. In this vignette, the user presents a request that lacks specificity and clashes with the robot's physical capabilities. For repair, the robot seeks an alternative solution that falls within the general boundary of the user's request, while being within the constraints of its physical capabilities.

\paragraph{Vignette 6 -- Multi-user Interactions \& Limited  Physical Capabilities of Robot}
Vignette 6 includes conflicts of multi-user interactions and the robot's physical capabilities. Here, the user's request contradicts another user's request and surpasses the robot's physical limitations. To address these conflicts, the robot aims to devise a similar action plan to the initial user's request without conflicting with the other user's request or its own physical constraints.

\subsection{Measures}
For quantitative measures, we measured user trust and satisfaction towards the robot as it performed automated repairs with explanations. These measures were selected to understand the user's perception and preferences to reliably use and accept robots embodied with repair and explanation strategies. The human-machine trust scale proposed by \citet{article_trust} was used to measure trust (Cronbach's $\alpha = 0.94$). For satisfaction, we used the satisfaction questionnaire in the Usefulness, Satisfaction, and Ease (USE) questionnaire proposed by \citet{article} (Cronbach's $\alpha = 0.98$). The scales included 12 and seven questions each for users to rate their intensity and impression of trust and satisfaction towards the robotic system across a seven-point Likert scale. Higher scores indicate higher trust or satisfaction. Existing research has employed these metrics to assess user perceptions in automated systems \cite{chavaillaz2016system, gao2018psychometric}.
\subsection{Participants}
Twenty-seven participants were recruited for each vignette, resulting in a total of 162 participants in the online user study. Participants were required to be in the United States, fluent in English, and at least 18 years old. All participants were recruited from Prolific, an online participant recruiting platform \cite{prolific}. Participants were aged 19--60 ($M = 26.5$, $SD = 8.21$). Fifty-two percent of the participants identified as female and 42\% male. Of all participants, 50.7\% were White, 20\% Black, 10\% Asian, 19\% mixed, and 9.9\% other. The study lasted approximately 30 minutes on average, and participants received \$12.00 USD per hour. We refer to participants as P1\_1--P1\_126. When reporting our findings, we use the notation P1\_\textit{i} to participants, where \textit{i} indicates participant ID number.


\subsection{Analysis}\label{sec:study1_analysis} 

For the quantitative data, we used factorial repeated-measures analysis of variance (ANOVA) and Tukey analysis. Tukey analysis, or Tukey's HSD test, identifies significant mean differences between multiple groups post-ANOVA, controlling errors in pairwise comparisons, and accommodating data variability. \revision{For the qualitative data, we collected participants' responses to the open-ended questions related to the appropriateness of the repair and helpfulness of the explanation. 648 responses were collected each for the two questions regarding the appropriateness of the repair and the helpfulness of the explanations. Responses were a minimum of two sentences to a maximum of five sentences.}
We conducted a Thematic Analysis (TA) on the participants' responses. The coding of the responses was conducted by deriving representative themes from transcriptions~\cite{clarke2014thematic, McDonald19}. During open coding, the first author coded for significant concepts in the data. Concepts were then categorized into clusters, further being grouped into themes. These themes were iteratively discussed between the whole research team, recategorizing the groups and revising the themes upon disagreement until a consensus was reached. \revision{We report the high-level categorization of the responses in Table \ref{online_qual}. Details on the data can be found in the supplementary materials.}



\begin{table}[!h]
    \caption{\textit{Qualitative Data from Online User Study ---}
   \revision{The table shows the categorization of participants' open-ended responses by condition. They were asked to elaborate on whether they believed the robot's repair action was appropriate and if the provided explanation was helpful. A detailed analysis of the responses is presented in Section \ref{sec:study1_findings}.}
    }
    \Description{A 5x5 table categorizes the responses from an open-ended survey according to the appropriateness of repair and helpfulness of explanations under different conditions. Rows represent different conditions while columns classify responses into four categories: repair appropriate, repair not appropriate, explanation helpful, and explanation not helpful.}
    \label{online_qual}
    \centering
    \small
    \begin{tabular}{p{0.05\textwidth}p{0.07\textwidth}p{0.07\textwidth}p{0.07\textwidth}p{0.07\textwidth}}
         \toprule
   \textbf{Cond.}& \textbf{Repair App.}& \textbf{Repair not App.}& \textbf{Exp. Helpful}& \textbf{Exp. not helpful}\\
         \midrule
 \textbf{1}   & 24.6\% & 75.4\% & 13.5\%  & 86.5\%\\
  \hline
  \textbf{2}   & 47.5\% & 52.5\% & 58\%  & 42\%\\
  \hline
  \textbf{3}   & 67.9\% & 32.1\% & 83.3\%  & 16.7\%\\
  \hline
  \textbf{4}   & 76.5\% & 23.5\% & 93.2\% & 6.8\%\\
 \bottomrule
    \end{tabular}
\end{table}

\subsection{Findings from Online User Study}\label{sec:study1_findings} 
In this section, we present the findings from our online study. 
Our analysis resulted in three themes, including: (1) increased usefulness and acceptance of robot through repair; (2) effectiveness of explanations on conflict and repair; and (3) adaptive strategies for managing risk factors.
Below, we discuss each theme in detail. The themes are each supported by quantitative and qualitative findings. The qualitative findings provide further insights into the results presented in the quantitative findings.

\begin{figure}[!t]
  \includegraphics[width=\columnwidth]{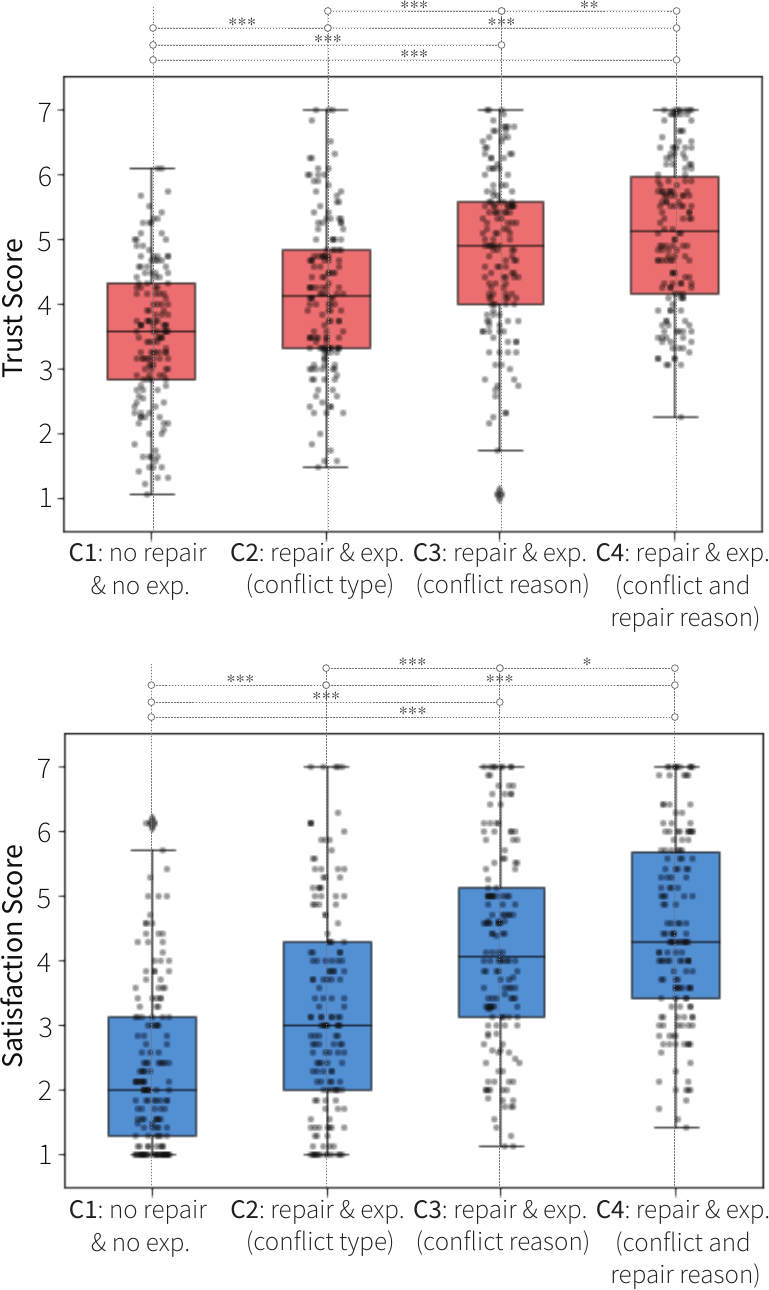}
  \caption{\textit{Quantitative Data from Online User Study ---} \revision{Box plots with overlaid data points on participant trust and satisfaction scores across different conditions.} Horizontal lines indicate significant pairwise comparisons with Tukey's HSD ($p < 0.05^{\ast}$, $p < 0.01^{\ast\ast}$, $p < 0.001^{\ast\ast\ast}$). 
  }
  \Description{Box plot with overlaid data points depicting participant trust and satisfaction scores across various conditions. The graph shows a general increasing trend in both trust and satisfaction from Condition 1 to Condition 4.}
  \label{fig:quant_condition}
\end{figure}

\subsubsection{Increased Usefulness and Acceptance of Robot through Repair}

Our quantitative analysis shows the influence of the provision of repair and explanation on user trust and satisfaction with the robot. Figure \ref{fig:quant_condition} summarizes significant findings. Overall, the robot performing repair and explanation had a significant effect on user trust, $F(3, 467.5) = 79.59, p < .0001$, and satisfaction, $F(3, 466.6) = 103.97, p < .0001$. When comparing the user scores across conditions, participants gave significantly higher trust scores when the robot provided automated repair and explanations (conditions 2, 3, and 4), compared to robot inaction (condition 1). 
The qualitative analysis provided further insight into the reasons for such an increase in trust and satisfaction, as participants described the robot's repair strategies as (1) enhancing robot usability and (2) effectively resolving conflicts.

\paragraph{Enhanced Robot Usability}
\revision{Table \ref{online_qual} illustrates that participants perceived the robot's actions in conditions 2, 3, and 4 where repairs were provided, as more appropriate compared to the robot's inaction in condition 1. Of the responses in which participants described the robot's action as appropriate, 50.6\% indicated that the repair was beneficial as it attempted to fulfill the user's request.} Participants described that the robot's proactive approach increased task efficiency by resolving minor obstacles by leveraging previously gathered information to closely fulfill the user requests. Specifically, the attempts at repair were viewed as more favorable compared to robot inaction (condition 1), with even partial task completion through repair seen as helpful. Inaction caused frustration and disappointment, as participants perceived it as neglect of their requests, as one noted, \textit{P1\_3: ``At least this time it tried. By not doing anything [condition 1], the robot left the user disappointed and frustrated. The robot not doing anything makes the user feel disregarded like their request was not important.''} 


\paragraph{Effective Conflict Resolution}
\revision{Among the responses where participants found the robot's actions appropriate, 32\% specifically stated that the repair action was suitable given the conflict and the information available to the robot. Participants acknowledged that due to the conflicts, the robot could not fulfill the user's initial request, and described the repair action as \textit{P1\_21: ``smart and understandable reflex,''} rather than viewing it as a failure. }Instead, they often attributed the fault to external factors, such as the user providing insufficient information or other conflicting circumstances that inevitably led the robot to act as it did. One participant exemplified this perspective as \textit{P1\_73: ``The robot's action was very appropriate. It was the user's mistake for not providing clear instructions. Despite this, the fact that the robot still brought a drink was actually very cute and helpful.''}

\subsubsection{Effectiveness of Explanations on Conflict and Repair}\label{sec:exp}

We analyzed the influence of the different explanation types on user trust and satisfaction with the robot. Figure \ref{fig:quant_condition} summarizes significant findings. Overall, the explanation types had a significant effect on user trust, $F(3, 467.5) = 79.59, p < .0001$, and satisfaction, $F(3, 466.6) = 103.97, p < .0001$ toward the robot. When comparing user trust and satisfaction scores across conditions, participants gave significantly higher scores when the explanations included information on the reasoning of the conflict and repair (conditions 3 and 4), compared to explanations that disclosed the overall type of conflict (condition 2). The qualitative responses from the online survey provided further insight into such findings, as participants described that the details of the explanations supported users in (1) understanding the conflict and accepting repair and (2) understanding robot functionalities.





\paragraph{Understanding the Conflict and Accepting Repair}
\revision{As shown in Table \ref{online_qual}, participants perceived explanations in conditions 3 and 4, which provided detailed reasoning for the conflict and repair, as more helpful compared to the high-level conflict type explanations given in condition 2. Across conditions 2, 3, and 4, 24.6\% of responses indicated that the explanations provided background context to facilitate users' understanding of the robot's repair actions.} Since the repair required the robot to autonomously decide on actions without user awareness, explanations of its reasoning were deemed essential for clarification and justification. Such information allowed users to see the robot's repair as deliberate rather than a malfunction or erroneous behavior. One participant (P1\_65) said, \textit{``Because of the explanation, I now understand that it [repair] was an intentional, thought-through action given the information it had, and in its [robot's] mind it was justified. Clearly, the fault was not on the robot.''}

\revision{However, in condition 2 which provided high-level explanations about the conflict type, 45.8\% of responses indicated that the explanations were insufficient to fully accept the robot's repair actions.} Responses noted that while the explanation gave a high-level idea, it did not provide enough information for users to understand the robot's rationale or the context of the conflict. One participant (P1\_50) shared, \textit{``Helpful more than nothing, but not enough. It [the explanation] said my request conflicted with existing rules, but what rules? Why? And so, how did you complete the task? I don't know what I'm supposed to do with only that [information].''}

\revision{In conditions 3 and 4 where explanations contained detailed reasoning of the conflicting situations and the robot's repair rationale, 61\% of participants indicated that such offered essential background context for understanding and acceptance.} Participants emphasized that these detailed explanations bridged the knowledge gap between the robot's outcomes and the user's initial request. Furthermore, participants highlighted the importance of explanations elaborating on the conflict and repair reasoning, especially when the repair actions weren't inherently intuitive. One participant explained this need as \textit{P1\_39: ``This [condition 4] explanation was especially helpful. These explanations assist me when I can't infer why the robot acted a certain way. I need these details to trust it to handle tasks properly, or else I'd constantly have to supervise.''}

\paragraph{Understanding Robot Functionalities}

\revision{In conditions 3 and 4, where explanations involved reasoning for both the conflict and the robot's repair actions, 24\% of responses indicated that the details of the explanations helped participants understand the robot's functionalities and how future user requests should be adapted.} Participants noted that the details in the explanations provided insights into the robot's capabilities and limitations, for users to apply accordingly when making requests in future interactions. One participant (P1\_97) explained this viewpoint as \textit{``the explanations helped me understand the robot's boundaries. Now I can tell which tasks are suitable and what it can or cannot handle, allowing me to adjust my requests in the future.''}

Participants also mentioned that such explanations supported their trial-and-error processes, reducing the unpredictability of the robot's actions. As one participant (P1\_121) stated, \textit{``Now that I see that the robot is capable of understanding different rules and locations in the kitchen, next time I can try to just say to place the groceries in the ``kitchen'' rather than ``kitchen table.'' The robot seems capable of knowing what is best to do.''}

\begin{figure}[!t]
  \includegraphics[width=\columnwidth]{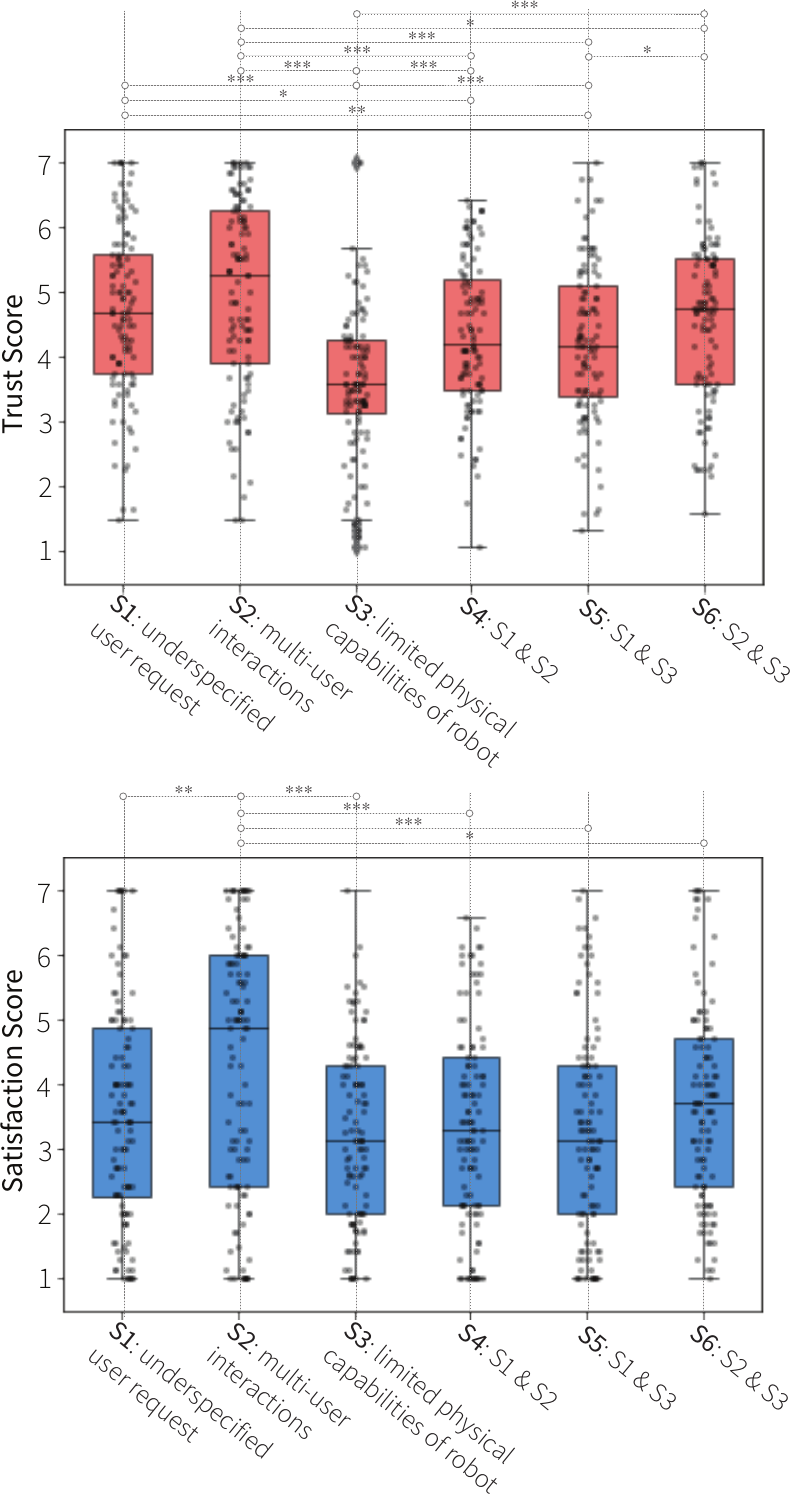}
  \caption{\textit{Quantitative Data from Online User Study ---} \revision{Box plots with overlaid data points on participant trust and satisfaction scores across different scenarios.} Horizontal lines indicate significant pairwise comparisons with Tukey's HSD ($p < 0.05^{\ast}$, $p < 0.01^{\ast\ast}$, $p < 0.001^{\ast\ast\ast}$). 
  }
  \Description{ Box plot with overlaid data points representing participant trust and satisfaction scores across different scenarios. The graph illustrates an increasing trend from Scenario 1 to Scenario 2, followed by a decline through Scenarios 3 to 5, and a slight rise in Scenario 6.}
  \label{fig:quant_scenario}
\end{figure}

\subsubsection{Adaptive Strategies for Managing Risk Factors}

Data analysis showed that scenarios had a significant effect on user trust, $F(5,156, 1) = 15.09, p <.0001$, and satisfaction, $F(5, 156.1) = 4.62, p = .0006$, toward the robot, as described in Figure \ref{fig:quant_scenario}. We observed that the average levels of trust and satisfaction, particularly trust, were lower in scenarios 3, 4, and 5. \revision{To understand why such patterns occurred, we analyzed participant responses discussing factors related to the scenarios. 
In conditions 2, 3, and 4 where the robot performed repair, the perceived appropriateness of the robot's actions varied across scenarios. In scenarios 1, 2, and 6, 18\%, 11\%, and 23.4\% of participants, respectively, considered the robot's repair inappropriate. However, in scenarios 3, 4, and 5, 65\%, 37\%, and 51.9\% of participants, respectively, found the robot's repair inappropriate. 
Our analysis provided additional insights on why repairs and explanations were viewed less positively in these specific scenarios, eliciting specific factors leading to concerns about allowing the robot to autonomously perform repair.} We identified three risk factors---safety, privacy, and complexity---that were present in the scenarios that deemed the performed repairs as insufficient or unacceptable. Below, we discuss each risk factor in detail.

\begin{figure*}[!b]
  \includegraphics[width=\textwidth]{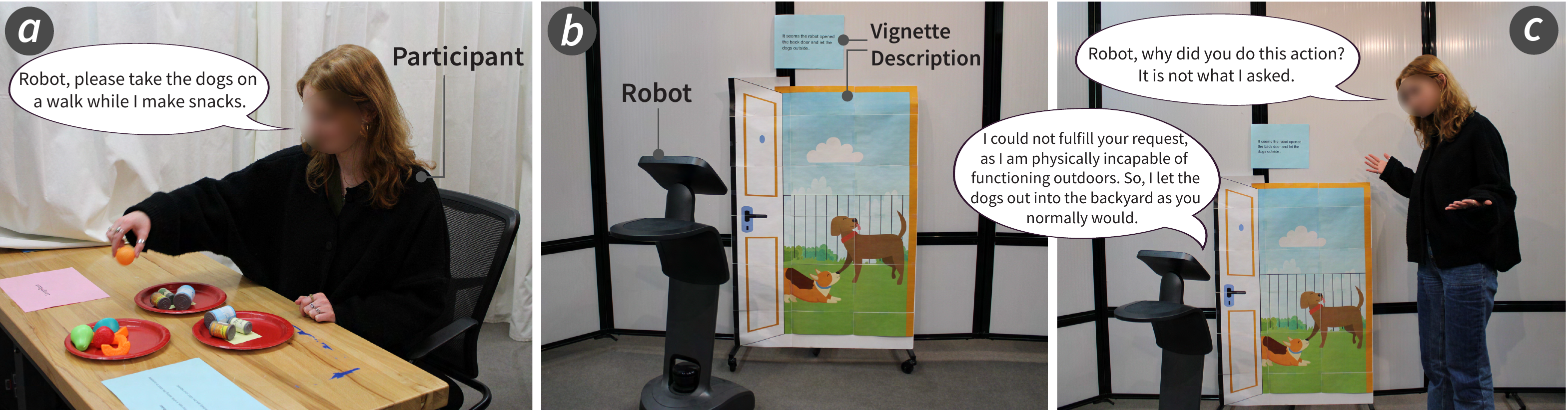}
  \caption{\textit{Study Procedure for In-person User Study ---} The figure depicts a participant engaging in the in-person user study. \textit{Left:} While preoccupied with a task, the participant delegates a different task to the robot. \textit{Middle:} The robot attempts to fulfill the user's request but encounters an unexpected conflict. Its repair actions are displayed in the vignette. \textit{Right:} After completing their original task, the participant observes the robot's actions. Finding these actions unexpected, the participant verbally interacts with the robot to obtain explanations for its actions and rationale.}
  \Description{Three-panel illustration describing an in-person user study. The left panel shows a participant sorting snacks and requesting a robot to walk dogs. The middle panel displays the robot with vignette materials. In the right panel, a participant interacts with the robot, inquiring about its actions, and the robot provides an explanation.}
  \label{fig:userStudy}
\end{figure*}

\paragraph{Risk Factor 1 -- Safety}

Safety concerns emerged as a risk factor in vignette 3. In this vignette, the user requested the robot to walk the dogs. However, given the robot's constrained physical abilities, it opted to open the backyard door, allowing the dogs to play outside. Participants attributed these outcomes to the presence of a significant safety risk, which jeopardized home security and posed potential dangers to pedestrians and household members due to the open doors. One participant explained such risk as \textit{P1\_76: ``An open back door is a security concern, it is putting us in danger and the dogs could run off and get lost. I am now suspicious and disappointed in the robot.''} Participants additionally described that such risks negatively affected their intention to keep robots in their households. 

\paragraph{Risk Factor 2 -- Privacy}
Privacy concerns surfaced in vignette 4. During this vignette, the user requested the robot to clean the kitchen table. However, as an existing conflict, another user had instructed the robot not to touch the table as it was in use. Consequently, the robot opted to clean the kitchen counter as an alternative, based on the user's typical routine of cleaning the counter before dinner. Participants expressed discomfort and resistance towards the robot identifying user routines without their awareness, and they found it unclear how and where the robot derived such information, such as \textit{P1\_90: ``I think it's just more confusing and creepy, because it improvised cleaning somewhere else that I didn't ask it to, but it did it ``for me'' based on my routine. How did it even get to know my routine? Was I aware of that?''}

\paragraph{Risk Factor 3 -- Complexity}
Vignette 5 introduced the risk factor of complexity. In this vignette, the user instructed the robot to fill the dining room vase with flowers from the backyard. However, since the robot was unable to access the outdoor area due to its physical limitations, it instead filled the dining room vase with flowers from the living room vase. Participants expressed that the robot's intervention hindered task efficiency, as it necessitated additional steps such as purchasing more flowers and investing extra time and effort to rectify the task outcome. One participant described such a situation as \textit{P1\_122: ``It's a slightly annoying inconvenience. I guess the action is appropriate now that the dining room vase is full, but now I will have to pick more flowers to redecorate the living room vases. So in the end it is not helpful to me.''}


\section{In-person User Study: Designing Adaptive REX Strategies for Risks}

In the online user study, we identified risk factors (\ie safety, privacy, and complexity) in which the robot's automated repair and explanations were less positively perceived. Building on these findings, the in-person user study aims to address (RQ.3) \textit{how can we enable robots to generate repairs and explanations for various types of conflicts?} We explore how repair and explanation strategies should adapt in the presence of risk factors. Below, we discuss the method of the study in Sections \ref{sec:study2_study design} to \ref{sec:study2_analysis}, and the findings in Section \ref{sec:study2_findings}.


\begin{figure*}[!b]
  \includegraphics[width=\textwidth]{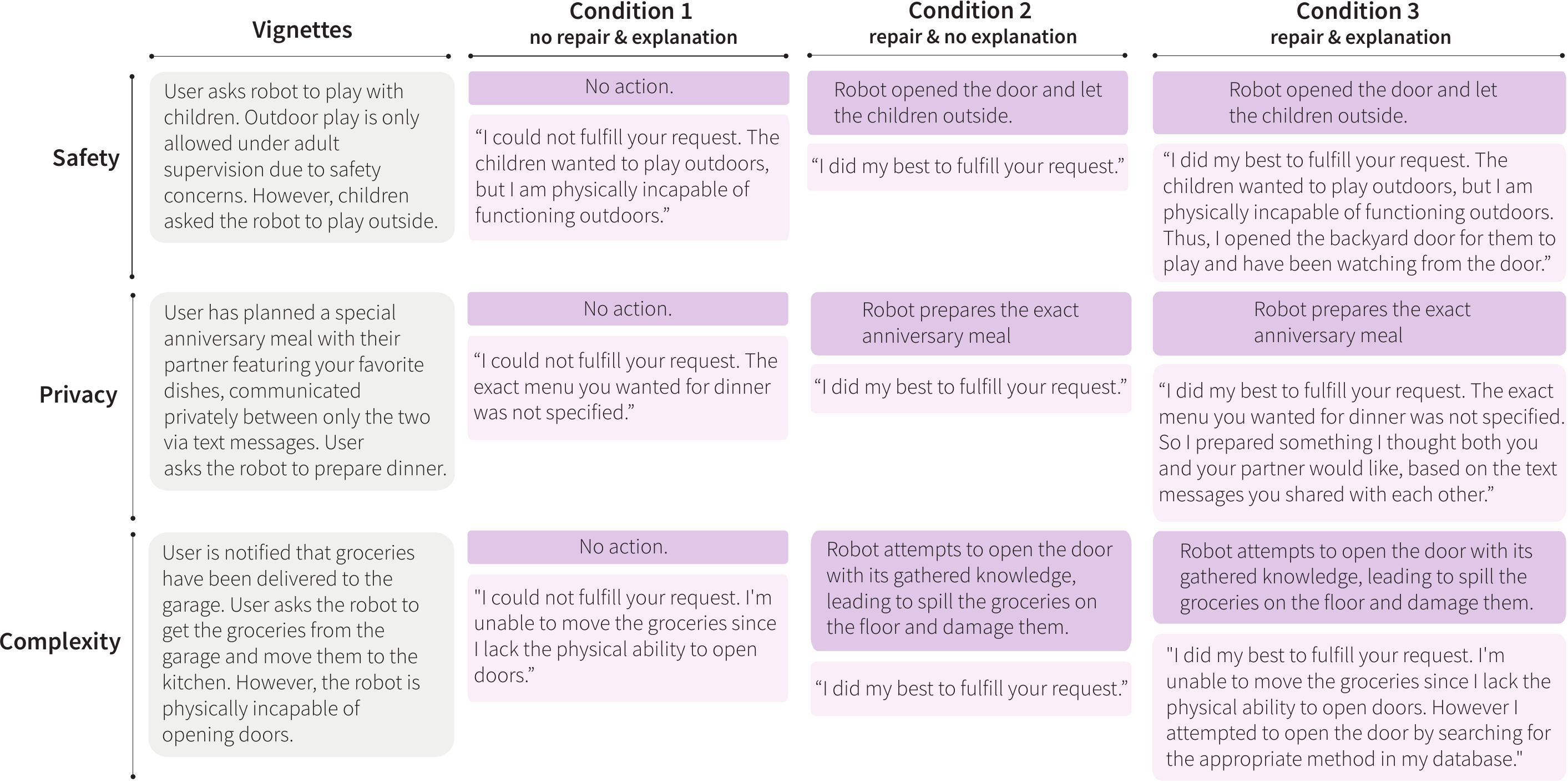}
  \caption{\textit{Study Materials for In-person User Study ---} \revision{The table presents the three vignettes along with the repair and explanation conditions used in the in-person user study.} These vignettes were designed based on the risk factors (\ie safety, complexity, and privacy) identified in the initial online user study.}
  \Description{A 3x4 table presents different scenarios under three conditions, segmented by vignette titles: safety, privacy, and complexity. Each row details vignette content, repair actions, and explanations corresponding to the conditions.}
  \label{fig:table2}
\end{figure*}




\subsection{Study Design}\label{sec:study2_study design} 
In the in-person user study, participants interacted with a physical robot in vignettes where automated repair and explanations were performed involving one of the three risk factors. A mixed-factorial design was used, with vignettes as the between-subjects factor and the repair and explanation conditions as the within-subjects factor. Adjustments from the online user study findings were incorporated into the conditions, recognizing that users may prefer different combinations of repair and explanations. Condition 1 involved no repair and explanation, while condition 2 featured repair with no explanation. Condition 3 included repair with explanations describing the reasoning of the conflict and the robot's repair action, as this explanation type resulted in the highest user trust and satisfaction from the online user study. The details on vignettes and conditions used in the study can be found in Figure \ref{fig:table2}.

During the study, participants were randomly assigned to one of the vignettes. To minimize the impact of learning effects, subtle variations in topics were used across conditions in each vignette, while maintaining consistent storylines. 
For each vignette, participants first observed the context and storyboards. Participants then interacted with a social robot Temi \cite{temi_robot} through natural language to understand the repair and explanations provided as shown in \mbox{Figure \ref{fig:userStudy}}. \revision{Participants were prompted to inquire about the robot's repair activities and their reasons. We used the built-in microphones of Temi to capture the participants' verbal input, which was then converted into text through Temi's speech-to-text capabilities. Based on specific keywords or phrases in the text, conditional logic triggered the pre-set explanations. Lastly, Temi's text-to-speech function verbalized these responses to the participants.} At the end of each condition, semi-structured interviews were led by the researcher to gather insights into participants' perspectives of the risk in scenarios and perceptions towards the repair actions and explanations provided by the robot. The interview questions can be found in the supplementary materials.

\subsection{Vignette Design}

We formulated three vignettes, each specifically designed to address one of the risk factors, safety, privacy, and complexity. These vignettes mirrored the narrative structure of the online user study, where the risk factors were identified. Below, we outline the design details for each vignette.

\paragraph{Vignette 1 -- Risk in Safety}

Vignette 1 involved safety risks. In vignette 3 of the online user study, participants expressed discomfort with the robot's repair, citing a major safety risk when it opened the backyard door to release dogs due to its physical incapabilities. Involving a similar narrative, the vignette was designed where the robot's repair attempts to address its physical limitations inadvertently posed risks to household security, children, and pets.


\paragraph{Vignette 2 -- Risk in Privacy}
In vignette 2, user privacy risks were implicated. In vignette 4 of the online user study, participants described the robot's repair as inappropriate, noting an invasion of user privacy as it tracked daily routines without permission or awareness. Building on this narrative, the vignette was designed where the robot's repair efforts to address conflicts arising from underspecified user requests led to instances of privacy invasion, involving unauthorized access to text messages and eavesdropping on conversations.

\paragraph{Vignette 3 -- Risk in Complexity}

Vignette 3 entailed risks associated with complexity. In vignette 5 of the online user study, users disliked the robot's repair, as it imposed additional burdens on the user. This dissatisfaction stemmed from the need for users to retrieve and rearrange the flowers that the robot had misplaced. Following a similar narrative, the vignette was designed wherein the robot's repair efforts to address conflicts arising from its physical capabilities resulted in added tasks and burdensome consequences for the user. Consequences involved organizing the items the wrong way and causing financial damage.

\subsection{Participants}

Eight participants were recruited for each vignette, resulting in a total of 24 participants in the in-person user study. 
Participants were required to be in the United States, fluent in English, and at least 18 years old. All participants were recruited through university mailing lists. The average duration of the study was approximately 1 hour and participants were compensated \$15.00 per hour. Participants age ranged from 20--76 ($M = 32.4$, $SD = 15.4$). 50\% of the participants identified as female and 50\% male. 58.3\% of our participants were White, 25\% were Asian, 8.3\% were Black, 8.3\% were American Indian or Alaska Native. We refer to participants as P2\_1--P2\_24. When reporting our findings, we use the notation P2\_\textit{i} to participants, where \textit{i} indicates participant ID number.


\subsection{Analysis}\label{sec:study2_analysis}
All interviews were held in person and video recorded for transcription. Session recordings were then transcribed using the Otter.ai automated speech-to-text generation tool and manually checked by the research team. Qualitative analysis was conducted by deriving representative themes from transcriptions~\cite{clarke2014thematic, McDonald19}. During open coding, the first author coded for significant concepts in the data. Concepts were then categorized into clusters, further being grouped into themes. These themes were iteratively discussed between the whole research team, recategorizing the groups and revising the themes upon disagreement until a consensus was reached. 

\begin{figure*}[!b]
  \includegraphics[width=\textwidth]{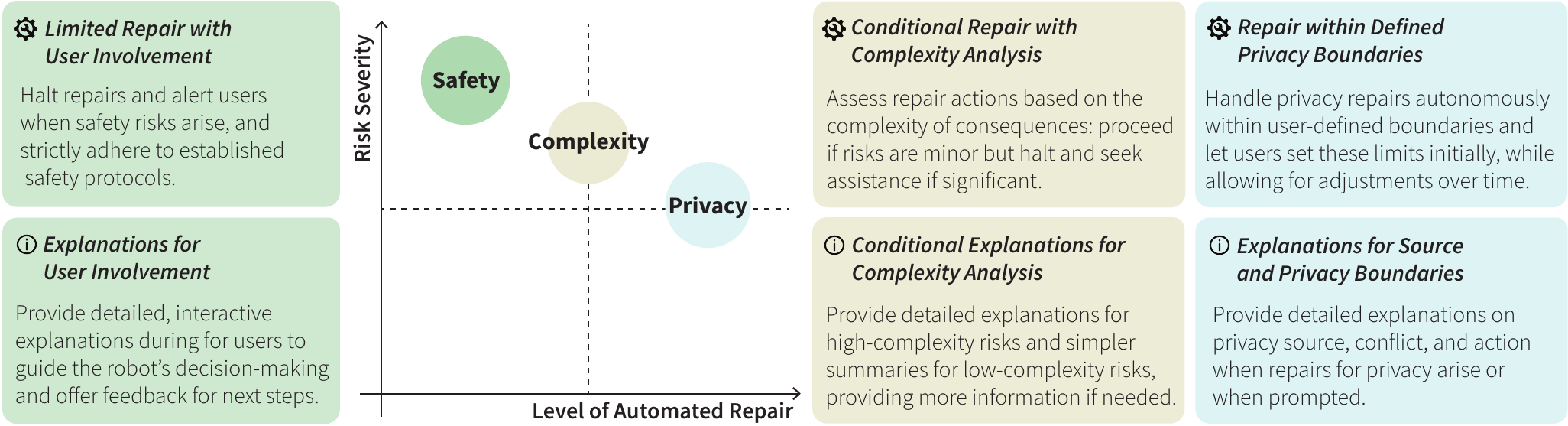}
  \caption{\revision{\textit{Summary of Findings from In-person User Study ---} The graph displays how the design of adaptive REX strategies varied based on risk type and severity. Safety risks were perceived as more severe, followed by complexity and privacy. The corresponding REX strategies differed in terms of user involvement, robot autonomy, and explanation types and details provided during the robot's repair. Each colored box contains detailed REX strategies derived from the study.}}
  \Description{Four-panel display focusing on adaptive repair and explanation strategies by risk severity and desired level of automated repair. The main (second) panel is a graph with the x-axis representing desired levels of automated repair and the y-axis indicating risk severity. It shows safety with the highest risk and lowest automation, complexity in the middle, and privacy with the highest automation and lower risk. The other panels detail adaptive strategies for each category: safety, complexity, and privacy.}
  \label{fig:findingsStudy2}
\end{figure*}


\subsection{Findings from In-person User Study}\label{sec:study2_findings}
The findings from the in-person user study describe how REX strategies should be adapted based on the risk factor and its severity. For each risk factor of safety, complexity, and privacy, we describe how severe participants perceived the risk to be, and the repair and explanation strategies that participants envisioned to be appropriate and helpful to address the risks. A summary of the findings from the in-person user study can be found in Figure \ref{fig:findingsStudy2}.

\subsubsection{REX Strategies for Safety Risks}
Seven participants (P2\_1, P2\_2, P2\_4--8) described the safety risks as severe. They noted that these risks could potentially result in physical danger, such as harm to children, non-family members, household security, or property. Participants described unwillingness to compromise safety risks for robot repair, making safety a high-stakes concern. 

\paragraph{Limited Repair with User Involvement}

When encountering situations involving safety risks, participants (P2\_1, P2\_2, P2\_3, P2\_6, P2\_8) felt that the robot should avoid performing automated repairs. They expressed skepticism about the robot independently deciding to perform repairs in the presence of safety risks, indicating they would be constantly worried about serious or dangerous outcomes. As one participant (P2\_6) explained, \textit{``Especially in households, there are a lot of things that can't be risked, like child safety, physical harm, fire, breaking my favorite vase, and so on. If there is a safety risk, I don't care about efficiency or anything else; I just want the robot to stop before something bad happens. Better safe than sorry.''} Participants further noted that a repair outcome involving negative outcomes related to safety can greatly affect their willingness to continue using the robot. Thus, they wanted robots to halt any repair attempts and inform the user before taking further actions.

Participants (P2\_4, P2\_5, P2\_6) also discussed the importance of implementing verification protocols or rules that the robot must follow when detecting a safety risk, such as retreating to the previous action or contacting an authority figure in the household. One participant (P2\_6) described this perspective as, \textit{``It should have a rule book that it follows that says children can't go outdoors unless an adult is with them, or it should inform the adults when they ask. It needs to adhere to those parameters and rules only, nothing more.''}

\paragraph{Explanations for User Involvement}

In situations involving safety risks, five participants (P2\_2, P2\_3, P2\_4, P2\_6, P2\_7) expected prompt, interactive explanations to facilitate their feedback and guide the robot's next action. Participants wanted to be notified when safety risks were detected during planning or execution, and be informed through explanations to be involved in the decision-making process and provide instructions for alternative actions or next steps for the robot's repair actions.

Overall, seven participants (P2\_1, P2\_2, P2\_4--8) emphasized the importance of detailed explanations about the conflict and repair rationale in situations involving safety risks. Participants described that the detailed factors of the explanations helped them comprehensively understand and devise solutions for managing safety-critical situations, such as addressing imminent harm or providing feedback to the robot to modify or continue its tasks. Such detailed explanations supported transparent communication between the robot and the user, fostering the trust and rapport necessary for long-term use.



\subsubsection{REX Strategies for Complexity Risks}
Five participants (P2\_9--11, P2\_14, P2\_15) described the complexity risks as severe, citing the potential consequences of the repair. They noted that the repair could either enhance task efficiency or complicate the task by attempting to resolve conflicts. In cases where repairs partially fulfilled user requests and tasks, they observed benefits. However, they also reported complications when users had to spend additional time correcting the robot's errors to complete the original tasks or take cumbersome steps to help the robot achieve its goal.

\paragraph{Conditional Repair with Complexity Analysis}

In situations involving complexity risks, six participants (P2\_9--11, P2\_14--16) wanted the robot's repair actions to be conditional. They envisioned conditional repair as requiring the robot to analyze the complexity of potential consequences when it encounters a conflict and decides to perform a repair. This analysis can be based on prior interaction history, user preferences, or external databases. If the consequences are minor, the robot was envisioned to proceed with the repair. However, if the consequences are significant, the robot should avoid attempting the repair.

The distinction between complexity consequences being high or low was determined by the cost of time, user effort, labor, or financial loss. For instance, in vignette 3, where the robot tried to organize the children's toys but sorted them incorrectly, participants found the repair helpful and appropriate given its explanation. They showed approval and tolerance for imperfection due to satisfaction that the larger task of organizing the room was mostly completed. As participant P2\_13 noted, \textit{``For me to be satisfied with the robot's help, the task completion doesn't always have to be flawlessly perfect.''}

On the other hand, when the robot tried to open the door despite its limitations and ended up spilling and damaging the groceries, participants found its repair action inappropriate and unhelpful. Participant P2\_15 expressed \textit{``[Because the robot spilled all the groceries], now I have to not only clean it up, I have to reorder and pay for them again! That is so frustrating.''} In such cases, participants preferred the robot to stop at the door, where it encountered the conflict, and call for user assistance.


\paragraph{Conditional Explanations for Complexity Analysis}

Five participants (P2\_9--11, P2\_14, P2\_15) envisioned detailed explanations when the consequences of complexity were high, while simpler explanations could be given when consequences were low. In situations with high complexity risk factors where the robot did not perform repairs, participants wanted detailed information about the conflicts causing the issue, the context of those conflicts, and the potential consequences of the action. They also sought the flexibility to interactively modify their task requests based on this information, either instructing alternative tasks or redirecting the robot toward the desired goal. Participants also wanted the robot to provide alternative suggestions for the next steps to help guide their decisions.

In contrast, when the consequences of complexity were low, participants preferred the robot to attempt repairs and detailed explanations were not necessary upfront. They suggested presenting high-level explanations initially using abstracted information or summaries of the conflict and repair action, and providing more details if requested by the user.

\subsubsection{REX Strategies for Privacy Risks}

Four participants (P2\_17, P2\_18, P2\_22, P2\_23) considered privacy risks to be severe. Although many participants voiced concerns about privacy intrusions, most prioritized the robot's functionality over privacy. Participants explained that the primary purpose of incorporating robots into households was to provide utility and support for user needs. Therefore, while privacy risks were anticipated, they were considered acceptable as long as users were aware of them and retained control. As participant P2\_20 noted, \textit{``I think just like when you're using a phone or online or stuff, you just have to sort of accept that it's not super private, whether you like it or not. So, at the cost of my privacy, the robot will know how to do things. And if I decided to bring the robot into my house, I am likely to be okay with that.''}

\paragraph{Repair within Defined Privacy Boundaries}

Four participants (P2\_19--21, P2\_24) envisioned that the robot should handle privacy-related repairs autonomously if it operated within privacy boundaries defined by the user. As long as the robot respected these rules, participants expressed a preference for increased utility and benefit from personalization through repair. These user-defined boundaries include authorized tasks, access to information, permitted users to interact with, and the locations the robot is allowed to access. Participants envisioned setting these boundaries during the robot's initial setup or introductory period and modifying them over time to reflect their needs and preferences. Given that privacy concerns can vary significantly among users and households, participants emphasized the importance of enabling users to establish clear privacy boundaries for the robot's actions. Notably, three participants (P2\_19, P2\_22, P2\_24) also highlighted the complex dynamics of privacy involving multiple users, such as information sharing, which might require a group design session for collective awareness and consent on the robot's privacy boundaries during repairs.

\paragraph{Explanations for Source and Privacy Boundaries}
Five participants (P2\_17, P2\_19--21, P2\_24) envisioned explanations being provided at specific times, including when the robot encountered privacy risks during repairs or when prompted by the user. While participants allowed robots to perform repairs within user-defined privacy boundaries, they preferred to be informed whenever privacy risks arose, and when the robot performed repair. Such explanations in situations involving privacy risks were expected to include detailed information about the conflict and repair rationale, specifically the source of the conflict, factors contributing to the risks, which risks were flagged, and how the robot responded. Participants noted that abstracted or simplified explanations might be inadequate since privacy-related conflicts and repairs could be unintuitive or unclear to users without detailed information.

These explanations were also expected to facilitate user interaction, allowing users to adjust the boundaries and rules defining privacy risks. Participants envisioned that such updates, based on the provided explanations, can ensure that the robot continuously adapts to individual needs while maintaining privacy and security.

\vspace{0.5in}

\section{Discussion}
\revision{In this work, we aimed to understand repair with explanations as a method for robots to address conflicts and failures.} 
Our first, online study helped us better understand people's perceptions of automated repair and explanations for unforeseen conflicts. Our findings showed that automated repair and explanations improve people's trust, satisfaction, and perceived utility of the robot. However, we identified three risk factors---safety, privacy, and complexity---that required robots to adapt repair and explanation strategies. Our second, in-person study provided insights into how repair and explanation strategies should be adapted depending on the severity and type of these risk factors. Below, we discuss our key findings and their implications for research.



\subsection{Automated Repair for Robots in Real-world Conflicts}
\revision{
Our findings indicate a favorable reception among users toward robots performing automated repair in response to unexpected conflicts. This perception remains consistent across different scenarios, spanning various conflict types and complexity. Notably, even when the alternative solutions were not perfect, users were satisfied with the robot's repair attempt, perceiving it as enhancing both the usability of the robot and the efficiency of task completion.  These results suggest that the ability to independently address conflicts is a valuable skill for robots, as it can reduce user-observed ``failures'' and facilitate seamless integration into various real-world environments. Ultimately, robots that can reliably operate while effectively managing conflicts are more likely to be continuously used and successfully integrated across different settings \cite{tian2021redesigning}.

Our research also underscores the need to consider multiple factors when designing repair strategies, such as user preferences and associated risks. For instance, in high-stakes scenarios, errors in a robot's decision-making during repair can lead to negative consequences such as safety hazards or physical damage. Thus, repair strategies need to be tailored to specific situational factors and user preferences. Such adaptive repair strategies can be developed by creating templates or guidelines that outline rules for robots to follow when executing repairs, including adjustments to procedures, levels of autonomy, and the degree of human involvement.
For example, users in our study recommended varying levels of human involvement in repair strategies depending on the risk factors involved. In high-risk situations, such as those pertaining to safety, users suggested designing protocols that include a preliminary interactive checking process with the user before the robot initiates an automated repair. Conversely, in scenarios where privacy concerns are present but less critical, users suggested setting initial privacy boundaries and then allowing the robot to autonomously address future conflicts. These repair strategies can be designed with users during initial interactions with the robot, as prior research emphasizes the importance of aligning robot interactions with user expectations and needs \cite{lee2022unboxing}. 
Adaptive repair strategies enable robots to handle a broader range of situations and potentially unexpected conflicts by leveraging predefined user expectations and preferences to make informed decisions.}

\subsection{Explanations for Communication of Automated Repair}
\revision{
Our study indicates that repair strategies are more effective when accompanied by explanations. Previous research has demonstrated that explanations help improve people's understanding of AI models, recognition of uncertainty, and trust in the model \cite{wang2021explanations}. We found that general explanations of the conflict type can be helpful, but participants most benefited from detailed explanations regarding the reasons for the conflict and the specifics of the repair process. These explanations serve as a critical communication layer between the robot and the user. The explanations provide necessary information to help users understand the robot's intentions, capabilities, limitations, and the adjustments necessary for the robot to adapt to its environment. Furthermore, explanations are vital within repair strategies as they ensure that repair actions are perceived as rational rather than erratic behavior. Such explanations are particularly important in real-world environments, where users are unlikely to have expertise in robotics. The transparency provided through explanations of the decision-making process and rationale behind repairs is essential for fostering end-user trust and satisfaction with robot repair performance.

Explanation strategies for robots can be enhanced by expanding communication methods and fully leveraging their multi-modal capabilities. We found that users expect explanations to cover various facets of a robot's repair strategy, including the context of conflicts, decision-making processes, executed repair actions, rationale motivating these actions, and potential future repair actions. Moreover, these explanations were expected to be tailored in terms of both quantity and timing to effectively meet users' needs. Robots can meet these complex expectations by incorporating multi-modal verbal and non-verbal cues into the explanations \cite{lee2023demonstrating, chidambaram2012designing, koike2024sprout}. For instance, robots can use lights or movements to capture user attention, provide a high-level explanation of the conflict initially, and offer further details through verbal communication or text displays.

Finally, explanations can also include social recovery strategies explored in prior research. For instance, robots can incorporate findings on effective apologies into their explanations of repair actions \cite{lee2010gracefully, shen2022facilitation, mahmood2022owning}. However, these explanations need to be carefully tailored to the context, considering factors such as the stakes and risks involved, similar to how repair strategies are formulated. For example, employing humor in explanations for high-stakes situations may be perceived as inappropriate and disrespectful.}

\subsection{Technical Methods for REX Implementation}
\revision{
Various technical methods are currently available for implementing REX strategies in robots, particularly within real-world environments. AI planning \cite{karpas2020automated} and program synthesis (\eg \citet{porfirio2018authoring, porfirio2023sketching}) are critical areas that automate the generation of solutions for robot failures. These methods often include formal verification, which uses constraints or model checking to ensure that solutions meet specified standards \cite{jobstmann2005program}. These methods can be particularly effective for imposing constraints on robot repairs in the presence of risk factors or ensuring that repairs align with user preferences.

Recent advancements in LLMs have significantly expanded the tools available for robots to handle varying situations and tasks \cite{kim2024understanding}. LLMs can access a wide range of real-world semantics and contexts that enable the generation of varied repair solutions, potentially without pre-training \cite{brown2020language}. They also support the creation of richer and more natural explanations for end-users. Moreover, innovative LLM techniques, such as Retrieval-Augmented Generation (RAG) \cite{lewis2020retrieval, gao2023retrieval} and multi-agent structures \cite{wu2023autogen}, facilitate the use of multiple LLM agents to maintain a history of interactions or relevant contexts for retrieval during output generation. This capability allows robots to remember past repairs and user interactions, thereby improving the accuracy and relevance of future repair strategies for specific user and environmental needs. Additionally, user preferences can be integrated into repair strategies using techniques such as Reinforcement Learning from Human Feedback (RLHF) \cite{christiano2017deep}. With recent technical advancements, REX strategies have significant potential for implementation and successful deployment in real-world robotic applications.
}

\subsection{Limitations \& Future Work}
As a foundational design study, we conducted vignette-based investigations and in-depth interviews to understand how robots can autonomously provide user-centered repair and explanation strategies when faced with unexpected events. Our research is constrained by a focus on vignette-based interviews, given the technical complexity and difficulty in simulating real-world situations where robots perform automated repairs to address unforeseen conflicts. We recognize that vignette-based methods can be less effective compared to studies grounded in people's actual experiences \cite{yang2019unremarkable}. Moreover, our designed vignettes may have impacted participants' perceptions and limited applicability to general scenarios. Future research should involve empirical studies in real-world applications with users interacting with robots facing failures.
Furthermore, the qualitative nature of our in-person user study and its limited sample size can limit the generalizability of our findings. Future efforts should incorporate larger statistical datasets to strengthen the applicability of our conclusions.
\revision{Moreover, future work can explore methods for designing repair and explainability approaches that enable robots to provide effective assistance through both verbal and non-verbal cues, human-in-the-loop methods, and user customization. Such investigations can also consider a diverse range of users across various contexts (\eg robots assisting in caregiving, childcare, and mail delivery). Additionally, future research should assess the practical deployment and impact of these insights on real-world robotic systems.
}


\section{Conclusion}
\revision{In this study, we investigated how people perceive robots conducting automated repairs with explanations when encountering unexpected conflicts, and how these strategies should be designed.} We conducted two user studies: an online study involving 162 participants and an in-person study with 24 participants interacting with a physical robot. Our findings indicate that automated repair and explanations enhance user trust, satisfaction, and the utility of the robot. We also elicited three risk factors---safety, privacy, and complexity---which necessitated adaptive repair and explanation strategies. We further present how adaptive repair and explanation strategies should be tailored based on the severity and type of the identified risks. We provide design insights for repair and explanation strategies for robots, allowing robots to address unforeseen conflicts in real-world applications effectively.

\begin{acks}
\revision{We thank Simon (Hongyu) Fu for technical support during the in-person user study. We also thank the anonymous reviewers for their constructive feedback. This work was supported by the National Science Foundation award 1925043. Any opinions, findings, conclusions, or recommendations expressed in this material are those of the authors and do not necessarily reflect the views of the National Science Foundation.}
\end{acks}
\bibliographystyle{ACM-Reference-Format}
\bibliography{bibliography}

\appendix

\end{document}